\pdfoutput=1

\documentclass[sigconf,natbib=true,anonymous=false]{acmart}

\usepackage{soul}
\usepackage{threeparttable}
\usepackage{multirow}
\usepackage{subcaption}
\usepackage{graphicx}
\usepackage[table,dvipsnames]{xcolor}
\usepackage{subcaption}

\usepackage{booktabs}
\usepackage{array}
\usepackage{fancyvrb}
\usepackage{makecell}
\usepackage{alltt}

\usepackage{enumitem}
\usepackage{xspace}
\usepackage{url}
\usepackage{csquotes}
\usepackage{amsmath}
\usepackage{hyperref}
\usepackage{listings}

\definecolor{CtxTwoK}{HTML}{D62728}      
\definecolor{CtxFourK}{HTML}{1F77B4}     
\definecolor{CtxEightK}{HTML}{2CA02C}    
\definecolor{CtxSixteenK}{HTML}{9467BD}  

\setcopyright{acmlicensed}
\copyrightyear{2025}
\acmYear{2025}
\acmDOI{XXXXXXX.XXXXXXX}

\acmConference[SIGIR '26]{The 49th International ACM SIGIR Conference on Research and Development in Information Retrieval}{July 20--24, 2026}{Melbourne, Australia}
\acmISBN{978-1-4503-XXXX-X/18/06}

\definecolor{ForestGreen}{rgb}{0.13, 0.55, 0.13}

\definecolor{VibrantBlue}{rgb}{0.0, 0.2, 1.0}

\setstcolor{VibrantBlue}

\begin{document}

\title{Context Shapes LLMs Retrieval-Augmented\\ Fact-Checking Effectiveness}

\author{Pietro Bernardelle}
\orcid{0009-0003-3657-9229} 
\affiliation{%
  \institution{The University of Queensland}
  \city{Brisbane}
  \country{Australia}
}
\email{p.bernardelle@uq.edu.au}

\author{Stefano Civelli}
\orcid{0009-0003-4982-9565} 
\affiliation{%
  \institution{The University of Queensland}
  \city{Brisbane}
  \country{Australia}
}
\email{s.civelli@uq.edu.au}

\author{Kevin Roitero}
\orcid{0000-0002-9191-3280} 
\affiliation{%
  \institution{University of Udine}
  \city{Udine}
  \country{Italy}
}
\email{kevin.roitero@uniud.it}

\author{Gianluca Demartini}
\orcid{0000-0002-7311-3693} 
\affiliation{%
  \institution{The University of Queensland}
  \city{Brisbane}
  \country{Australia}
}
\email{demartini@acm.org}

\renewcommand{\shortauthors}{Bernardelle et al.}

\renewcommand{\shorttitle}{Context Shapes LLMs Retrieval-Augmented Fact-Checking Effectiveness}
\begin{abstract}
Large language models (LLMs) show strong reasoning abilities across diverse tasks, yet their performance on extended contexts remains inconsistent. While prior research has emphasized mid-context degradation in question answering, this study examines the impact of context in LLM-based fact verification. Using three datasets (HOVER, FEVEROUS, and ClimateFEVER) and five open-source models accross different parameters sizes (7B, 32B and 70B parameters) and model families (Llama-3.1, Qwen2.5 and Qwen3), we evaluate both parametric factual knowledge and the impact of evidence placement across varying context lengths. We find that LLMs exhibit non-trivial parametric knowledge of factual claims and that their verification accuracy generally declines as context length increases. Similarly to what has been shown in previous works, in-context evidence placement plays a critical role with accuracy being consistently higher when relevant evidence appears near the beginning or end of the prompt and lower when placed mid-context. These results underscore the importance of prompt structure in retrieval-augmented fact-checking systems.
\end{abstract}
\maketitle
\section{Introduction}
\label{sec:intro}
The proliferation of misinformation poses a critical societal challenge, undermining democratic processes, public health, and social trust~\cite{Allcott2017,vosoughi2018spread}. Global events such as COVID-19 and recent political elections have highlighted the consequences of unchecked claims~\cite{Brennen2020,Garett2021,Guess2020}. The rapid spread of false information across digital platforms has far outpaced traditional, manual fact-checking, which remains time-consuming and resource-intensive~\cite{Hassan2017,Hassan2015}. This gap has driven growing interest in automated fact-checking systems capable of verifying claims at scale~\cite{Graves2018,thorne-vlachos-2018-automated}, with potential integration into social media and news platforms~\cite{Graves2018,Nakov2021,shaar-etal-2020-known,Shu2017}. Research initiatives such as FEVER and CheckThat! lab have further advanced this area~\cite{thorne-etal-2018-fever,barron2020checkthat}.

Advances in natural language processing and the emergence of large language models (LLMs) have opened new opportunities for automated verification. Their ability to process and generate human-like text, along with extensive knowledge assimilated during pre-trained, makes them strong candidates for claim verification~\cite{Singhal2023}. However, challenges remain: LLMs struggle with long contexts~\cite{Beltagy2020} and exhibit performance degradation when accessing mid-context information~\cite{liu-etal-2024-lost}. Unlike question answering, fact verification often requires synthesizing and reconciling conflicting information, further increasing task complexity.

In this paper, we investigate how LLMs handle extended contexts in claim verification. We first establish a parametric-only baseline, and then analyze how contextual factors shape performance. Specifically, we answer the following research questions: 
\begin{enumerate}[label=RQ\arabic*]
    \item \label{rq:pre} To what extent can LLMs verify claims using parametric knowledge alone, in the absence of external evidence?
    \item
    \label{rq:len}
    How does context length (i.e., the amount of context the model must process) affect LLM fact-checking accuracy?
    \item \label{rq:position} How does the position of relevant evidence within the prompt affect LLM fact-checking accuracy?
\end{enumerate}

Our results reveal that while models can resolve many claims from parametric knowledge alone, their performance is sensitive to how additional information is presented. Increasing input length introduces measurable degradation, and the relative position of evidence within the prompt substantially shapes outcomes. These findings indicate that effective fact verification depends not only on access to evidence, but also on how that evidence is organized within the model’s input.

\section{Background and Related Work}
In this section we examine two areas of relevant research: automated fact-checking systems and the evolution of verdict generation models ($\S$\ref{sec:bg:fact}), and the expanding capabilities of LLMs in processing long contexts ($\S$\ref{sec:bg:llms}).
\subsection{Automated Fact-Checking}
\label{sec:bg:fact}
Automated fact-checking has gained increasing attention due to the rapid spread of misinformation and the slow pace of manual verification~\cite{Adair2017,Hassan2015}. Early work \cite{vlachos-riedel-2014-fact} established a three-step framework—claim identification, evidence retrieval, and verdict generation—which continues to guide research in the field. While significant advances have been made in identifying claims and retrieving evidence~\cite{aly-etal-2021-fact,diggelmann2020climate,jiang-etal-2020-hover,kim-etal-2023-factkg,thorne-etal-2018-fever}, verdict generation remains challenging as it requires complex reasoning and the integration of multiple evidence sources. Building on this, the present work explores how LLMs can contribute to automated fact-checking, focusing on their capacity to use prior knowledge and manage long contexts to support verdict generation.

The evolution of verdict generation in automated fact-checking has progressed from rule-based systems to deep learning and, more recently, LLMs. Early neural approaches such as DeClarE~\cite{popat-etal-2018-declare} used bidirectional LSTMs and attention to jointly analyze claims and evidence, while transformer-based models like BERT~\cite{Soleimani2020ECIR} markedly improved claim-evidence reasoning. Subsequent frameworks, including GEAR~\cite{zhou-etal-2019-gear} and KGAT~\cite{liu-etal-2020-fine}, further enhanced multi-evidence reasoning.

The advent of LLMs introduced new opportunities and challenges. Although they exhibit strong reasoning abilities, they are prone to factual hallucinations~\cite{augenstein2024factuality}, prompting efforts toward retrieval-augmented architectures and improved evaluation methods~\cite{wang-etal-2024-factuality}. Recent work has also explored reducing reliance on external retrieval by distilling retrieval signals into model parameters~\cite{parametricrag2025}, highlighting ongoing trade-offs between parametric knowledge and explicit evidence grounding.

Our work contributes to this direction by examining how LLM reasoning can be effectively guided and constrained to produce reliable verdicts while leveraging their capacity for complex inference.
\subsection{Context Processing in LLMs.}
\label{sec:bg:llms}
Recent advancements in language model architectures and training have produced LLMs with vastly expanded context windows, reaching up to 200{,}000 tokens or more \cite{qwen2025qwen25technicalreport,yang2025qwen3technicalreport,dubey2024llama,geminiteam2024gemini15unlockingmultimodal,comanici2025gemini25pushingfrontier,openai2024gpt4technicalreport,cai2024internlm2technicalreport}. While these longer contexts enable richer reasoning over extended text, they also pose challenges in how effectively models utilize information across different positions. Prior work shows that LLMs often struggle to maintain performance when accessing mid-context information, performing best when relevant content appears near the beginning or end of the input \cite{liu-etal-2024-lost}. This ``lost in the middle'' effect echoes earlier findings that language models make increasingly coarse use of distant context \cite{khandelwal-etal-2018-sharp}, and persists even in advanced systems \cite{ivgi-etal-2023-efficient,krishna-etal-2022-rankgen}.
Evaluating long-context capabilities remains an open problem, as existing benchmarks mainly target retrieval rather than reasoning tasks \cite{hsieh2024,kamradt2023}. Recent efforts such as the Latent Structure Queries framework aim to assess models’ ability to synthesize information across long inputs \cite{vodrahalli2024}. 

In this work, we extend this line of inquiry to the domain of fact verification. We examine how context length and evidence positioning influence model performance in automated fact-checking, where integrating multiple pieces of evidence is essential for reliable verdict generation.
\section{Methodology}
This section describes the datasets used ($\S$\ref{sec:mt:data_models}), language models investigated ($\S$\ref{sec:mt:llms}), and the experimental procedures ($\S$\ref{sec:mt:exp}) designed to analyze how evidence placement and context length influence large language models’ fact-checking behavior.\footnote{Code and experimental artifacts will be made publicly available upon acceptance.}
\subsection{Datasets}
\label{sec:mt:data_models}
We conduct experiments on three claim verification datasets commonly used in the literature: HOVER~\cite{jiang-etal-2020-hover}, FEVEROUS~\cite{aly-etal-2021-fact}, and ClimateFEVER~\cite{diggelmann2020climate}. Each dataset consists of a collection of claims paired with a set of evidence sentences retrieved from Wikipedia (or related sources), along with a ground-truth label indicating whether the claim is supported or refuted by the provided evidence. To try minimizing confounds due to memorization we use the \emph{development set} of Hover and FEVEROUS, and the only available \emph{training set} of ClimateFEVER. Since label taxonomies differ across datasets, we unify them into a binary formulation with two labels: \texttt{SUPPORTS} and \texttt{REFUTES}. For comparability, we subsample FEVEROUS from its original 7,389 instances to 4,000 to match the size of HOVER, using label-balanced sampling. ClimateFEVER contains 907 instances and is therefore used in full.
\subsection{Language Models}
\label{sec:mt:llms}
We evaluate a range of large language model backbones spanning multiple families and scales, including LLaMA-3.1 (8B, 70B)~\cite{dubey2024llama}, Qwen-2.5 (7B)~\cite{qwen2025qwen25technicalreport}, and the more recent Qwen3 (8B, 32B)~\cite{yang2025qwen3technicalreport}. For all models, we use instruction-tuned variants~\cite{ouyang2022training}, which are better aligned with our in-context prompting setup for claim verification.
It is important to acknowledge that, like all LLMs, our chosen models may exhibit some degree of data leakage. While this overlap could enable models to recall specific information, it is worth noting that identifying and quantifying such leakage remains a significant challenge \cite{zhang2023counterfactual,zheng2021does,roberts2020much}. To address this concern, we conduct a baseline experiment that assesses the models' performance based solely on their parametric knowledge.

Although recent datasets have emerged~\cite{schlichtkrull2023averitec}, these resources still predate our models' training cutoff dates, potentially introducing similar data leakage concerns. Therefore, we opted to utilize more established benchmarks for our experimental evaluation.
At the same time, no instruct-based models were available before the release of these datasets, making it difficult to assess whether pretraining data contains explicit references to the evaluated claims. 
%
\subsection{Experimental Setting}
\label{sec:mt:exp}
We design a set of controlled experiments to isolate the effects of parametric knowledge, evidence availability, context length, and evidence placement on claim verification performance.

We begin by evaluating each model’s parametric knowledge (\ref{rq:pre}). In this setting, models receive only the claim text, without any retrieved evidence or external context, and must classify the claim as \texttt{SUPPORTS} or \texttt{REFUTES}. This condition serves as a baseline sanity check: it estimates how much models rely on memorized or internally stored knowledge and provides a reference point against which all evidence-based configurations are compared.

We then introduce retrieved evidence as in-context information. All evidence sentences associated with a claim are concatenated into a single evidence block and provided alongside the claim. This allows us to quantify the overall benefit of explicit evidence access relative to the parametric-only condition.

Finally, to systematically analyze contextual effects, we manipulate two variables: (i) total context length (\ref{rq:len}) and (ii) the relative position of the evidence block within the prompt (\ref{rq:position}).
The concatenated evidence block is inserted at 11 evenly spaced positions ranging from 0\% (beginning of the context) to 100\% (end of the context). To control for input size effects, we construct contexts of 2,048, 4,096, 8,192, and 16,384. In this setting, the evidence block is embedded within unrelated filler text\footnote{The filler text consists of a dummy passage sourced from \url{https://en.wikipedia.org/wiki/Llama}, used solely to control for context length while preserving evidence content.}, ensuring that its relative depth drives any performance differences.
We evaluate all combinations of the four context lengths and eleven evidence positions for each model, resulting in 44 distinct configurations per model. 
This setup enables us to disentangle (i) whether longer contexts degrade verification accuracy (\ref{rq:len}) and (ii) whether evidence effectiveness depends on its relative position within the prompt (\ref{rq:position}).
\section{Results}
\begin{table}
\centering
\caption{Parametric-only claim verification performance and accuracy when the gold evidence set is provided (w/ Ev.) without any addtional context. Gains are shown in parentheses.}

\label{tab:baseline_dataset_breakdown}
\scriptsize
\setlength{\tabcolsep}{2.5pt}
\begin{tabular}{llcccccccc}
\toprule
\textbf{Model} & \textbf{Dataset} & \textbf{Acc.} & \textbf{F1} 
& \multicolumn{2}{c}{\textbf{Precision}} 
& \multicolumn{2}{c}{\textbf{Recall}} 
& \makecell{\textbf{Acc.}\\\textbf{(w/ Ev.)}} \\
\cmidrule(lr){5-6} \cmidrule(lr){7-8}
 &  &  &  & \textbf{SUP} & \textbf{REF} & \textbf{SUP} & \textbf{REF} & \\
\midrule
\midrule
\multirow{3}{*}{Llama-3.1-8B}
& Hover & 0.55 & 0.55 & 0.47 & 0.64 & 0.57 & 0.54 & 0.61 (+.06) \\
 & FEVEROUS & 0.62 & 0.61 & 0.60 & 0.63 & 0.55 & 0.67 & 0.76 (+.14) \\
 & ClimateFever & 0.72 & 0.71 & 0.50 & 0.92 & 0.86 & 0.67 & 0.80 (+.08) \\
\midrule
\multirow{3}{*}{Qwen2.5-7B}& Hover & 0.53 & 0.53 & 0.46 & 0.69 & 0.75 & 0.38 & 0.63 (+.10) \\
 & FEVEROUS & 0.62 & 0.62 & 0.60 & 0.64 & 0.58 & 0.65 & \textbf{0.82} (+.20) \\
 & ClimateFever & 0.75 & 0.72 & 0.53 & 0.90 & 0.79 & 0.73 & 0.79 (+.04) \\
\midrule
\multirow{3}{*}{Qwen3-8B} & Hover & 0.61 & 0.58 & 0.54 & 0.64 & 0.40 & 0.76 & 0.63 (+.02) \\
 & FEVEROUS & 0.61 & 0.60 & 0.64 & 0.60 & 0.42 & 0.79 & 0.81 (+.20) \\
 & ClimateFever & 0.79 & 0.74 & 0.61 & 0.86 & 0.63 & 0.84 & 0.83 (+.04)
 \\
 
\midrule
\midrule

\multirow{3}{*}{Qwen3-32B} & Hover & \textbf{0.64} & 0.59 & 0.60 & 0.65 & 0.35 & 0.83 & \textbf{0.65} (+.01) \\
 & FEVEROUS & \textbf{0.64} & 0.62 & 0.65 & 0.63 & 0.48 & 0.77 & \textbf{0.82} (+.18) \\
 & ClimateFever & 0.82 & 0.77 & 0.68 & 0.87 & 0.67 & 0.88 & \textbf{0.85} (+.03) \\
\midrule
\multirow{3}{*}{Llama-3.1-70B} & Hover & 0.58 & 0.55 & 0.70 & 0.55 & 0.29 & 0.88 & 0.63 (+.05) \\
 & FEVEROUS & 0.62 & 0.59 & 0.78 & 0.58 & 0.34 & 0.91 & \textbf{0.82} (+.20) \\
 & ClimateFever & \textbf{0.83} & 0.78 & 0.73 & 0.86 & 0.63 & 0.91 & 0.84 (+.01) \\
\bottomrule
\end{tabular}
\end{table}
\subsection{Parametric Knowledge and Evidence Gains}
\label{sec:parametric-verification}
We first evaluate models in the parametric-only setting, where predictions rely solely on the claim text (\ref{rq:pre}). Table~\ref{tab:baseline_dataset_breakdown} reports accuracy, macro-F1, and label-wise metrics, together with the claim+evidence condition (without filler context).
Across datasets, all models perform above chance, indicating that a substantial portion of claims can be resolved using parametric knowledge alone. Performance varies more across datasets than across models. ClimateFEVER consistently yields the highest parametric scores, followed by FEVEROUS, while HOVER remains the most challenging benchmark.

Introducing gold evidence without additional filler yields systematic improvements across all models. Gains are modest for HOVER, moderate for ClimateFEVER, and largest for FEVEROUS (up to +.20 accuracy). These improvements confirm that explicit grounding provides complementary information beyond parametric knowledge, and the claim+evidence configuration serves as a reference upper bound before introducing longer contexts and positional manipulations. Additionally, as highlighted by the bold values in Table~\ref{tab:baseline_dataset_breakdown}, Qwen3-32B achieves the strongest accuracy on nearly every dataset under both the claim-only and claim+evidence conditions.

\subsection{The Effect of Context Length}
\begin{figure}
    \centering
    \includegraphics[width=\linewidth]{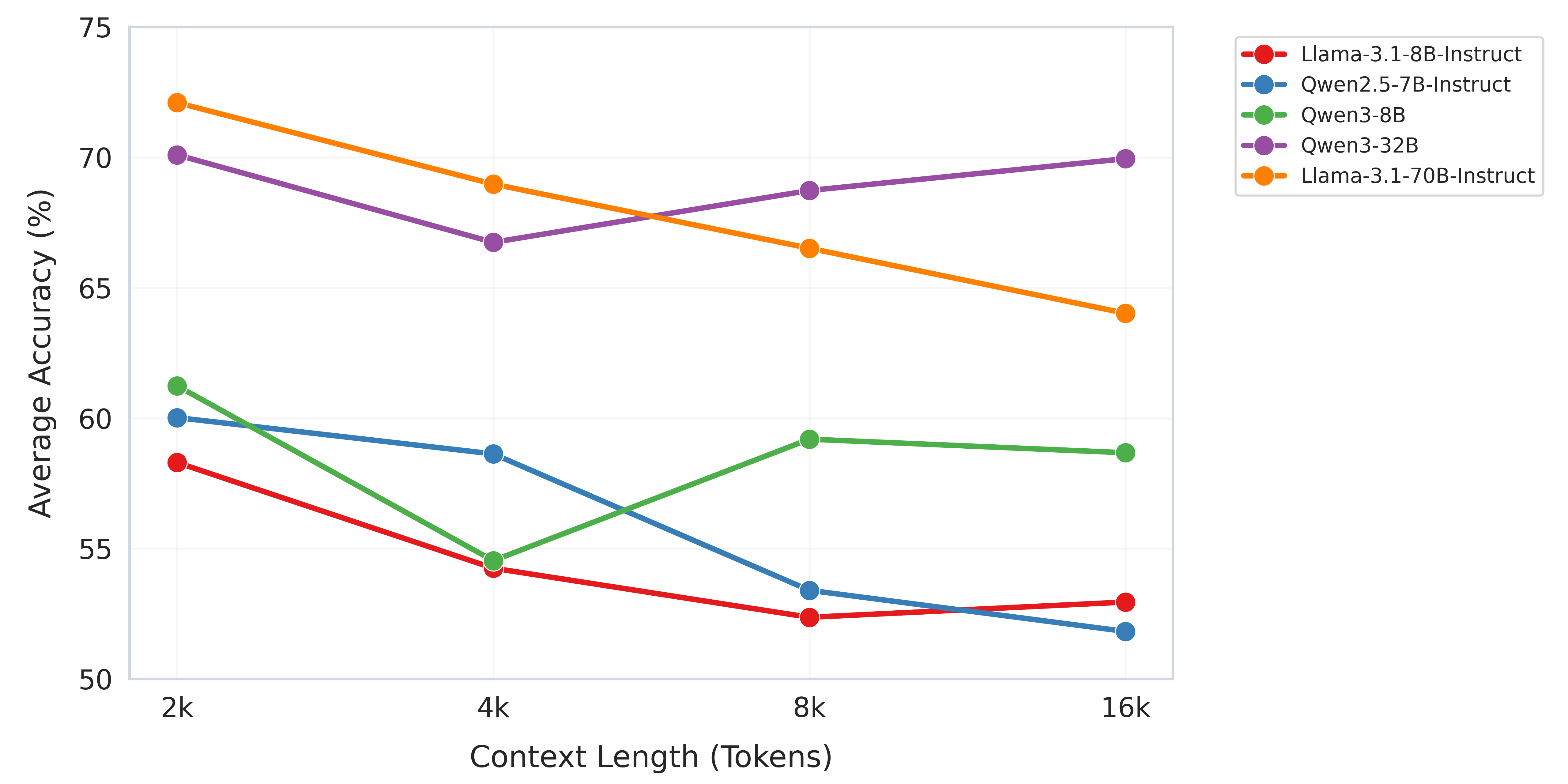}
    \caption{Accuracy as a function of context length averaged over evidence-placement runs.}
    \label{fig:avg_accuracy_vs_context_all}
\end{figure}
We next examine how context length influences verification accuracy (\ref{rq:len}). Figure~\ref{fig:avg_accuracy_vs_context_all} reports accuracy averaged over all evidence placements (0--100\%) for context sizes of 2k, 4k, 8k, and 16k tokens.

Across models, we observe a general decreasing trend as context length increases. While the magnitude of degradation varies, average performance at 8k and 16k is typically lower than at 2k, indicating that longer inputs introduce interference that reduces effective evidence utilization.
A notable exception emerges for both Qwen3 models (8B and 32B), which exhibit a distinct local minimum at 4k tokens before partially recovering at 8k. This non-monotonic pattern suggests that bigger context does not degrade performance linearly and may interact with internal attention allocation or training regimes at specific sequence lengths.
Among all evaluated systems, Qwen3-32B is the most stable across context lengths. Its performance curve in Figure~\ref{fig:avg_accuracy_vs_context_all} shows the smallest variation between minimum and maximum average accuracy. 

Overall, increasing context length tends to reduce fact-checking accuracy on average, even when gold evidence is present.

\subsection{Evidence Placement Effects}
\begin{figure*}
    \centering
    \includegraphics[width=1\linewidth]{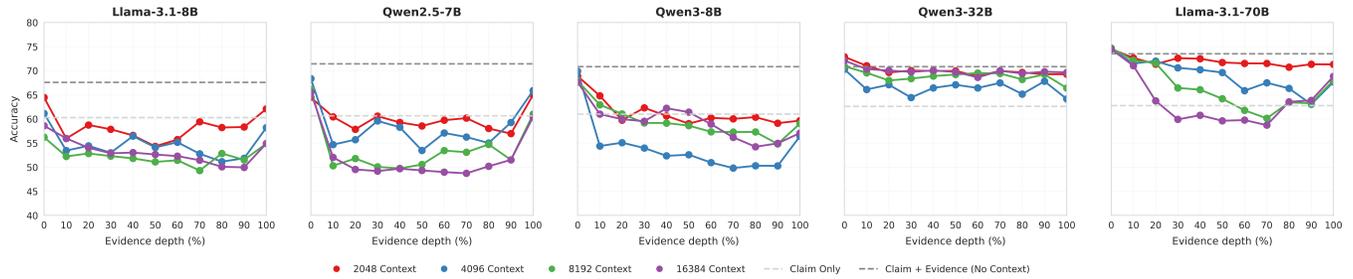}
    \caption{Accuracy vs.\ evidence depth (0--100\%) for the five investigated LLMs across four context lengths (2k/4k/8k/16k). Each panel shows how verification accuracy changes as the evidence block is moved through the prompt while total input length is held constant using filler text; horizontal baselines mark parametric-only (claim only) and claim+evidence without filler.}
    \label{fig:combined_accuracy_5models}
\end{figure*}
\begin{table*}
\caption{Sensitivity to context length by evidence position. For each model, dataset, and evidence depth (0--100\%), we report the best and worst accuracy observed across context lengths as \emph{Acc(BestCtx) / Acc(WorstCtx)}, with the corresponding context length shown in parentheses. Cell shading encodes $\Delta=\text{Best}-\text{Worst}$ (darker = larger degradation across context lengths).}
\label{tab:position-breakdown}
\tiny
\setlength{\tabcolsep}{1.1pt}
\begin{tabular}{llccccccccccc}
\toprule
\textbf{Model} & \textbf{Dataset} & \textbf{0} & \textbf{10} & \textbf{20} & \textbf{30} & \textbf{40} & \textbf{50} & \textbf{60} & \textbf{70} & \textbf{80} & \textbf{90} & \textbf{100} \\

\midrule
\midrule

\multirow{3}{*}{Llama-3.1-8B} & Hover
& \cellcolor{blue!15}0.58{\textcolor{CtxTwoK}{(2k)}} / 0.53{\textcolor{CtxEightK}{(8k)}}
& 0.54{\textcolor{CtxTwoK}{(2k)}} / 0.53{\textcolor{CtxFourK}{(4k)}}
& \cellcolor{blue!15}0.56{\textcolor{CtxTwoK}{(2k)}} / 0.51{\textcolor{CtxSixteenK}{(16k)}}
& \cellcolor{blue!8}0.55{\textcolor{CtxTwoK}{(2k)}} / 0.52{\textcolor{CtxSixteenK}{(16k)}}
& \cellcolor{blue!8}0.55{\textcolor{CtxTwoK}{(2k)}} / 0.52{\textcolor{CtxEightK}{(8k)}}
& \cellcolor{blue!8}0.54{\textcolor{CtxFourK}{(4k)}} / 0.52{\textcolor{CtxSixteenK}{(16k)}}
& \cellcolor{blue!8}0.54{\textcolor{CtxTwoK}{(2k)}} / 0.52{\textcolor{CtxSixteenK}{(16k)}}
& \cellcolor{blue!8}0.56{\textcolor{CtxTwoK}{(2k)}} / 0.52{\textcolor{CtxEightK}{(8k)}}
& \cellcolor{blue!8}0.56{\textcolor{CtxTwoK}{(2k)}} / 0.52{\textcolor{CtxSixteenK}{(16k)}}
& \cellcolor{blue!8}0.55{\textcolor{CtxTwoK}{(2k)}} / 0.52{\textcolor{CtxSixteenK}{(16k)}}
& \cellcolor{blue!8}0.55{\textcolor{CtxTwoK}{(2k)}} / 0.53{\textcolor{CtxSixteenK}{(16k)}} \\

& FEVEROUS
& \cellcolor{blue!25}0.70{\textcolor{CtxTwoK}{(2k)}} / 0.56{\textcolor{CtxEightK}{(8k)}}
& \cellcolor{blue!15}0.57{\textcolor{CtxSixteenK}{(16k)}} / 0.50{\textcolor{CtxEightK}{(8k)}}
& \cellcolor{blue!25}0.59{\textcolor{CtxTwoK}{(2k)}} / 0.49{\textcolor{CtxEightK}{(8k)}}
& \cellcolor{blue!15}0.58{\textcolor{CtxTwoK}{(2k)}} / 0.49{\textcolor{CtxEightK}{(8k)}}
& \cellcolor{blue!15}0.57{\textcolor{CtxTwoK}{(2k)}} / 0.49{\textcolor{CtxEightK}{(8k)}}
& \cellcolor{blue!15}0.53{\textcolor{CtxFourK}{(4k)}} / 0.46{\textcolor{CtxEightK}{(8k)}}
& \cellcolor{blue!15}0.57{\textcolor{CtxFourK}{(4k)}} / 0.49{\textcolor{CtxEightK}{(8k)}}
& \cellcolor{blue!25}0.60{\textcolor{CtxTwoK}{(2k)}} / 0.46{\textcolor{CtxEightK}{(8k)}}
& \cellcolor{blue!25}0.59{\textcolor{CtxTwoK}{(2k)}} / 0.47{\textcolor{CtxSixteenK}{(16k)}}
& \cellcolor{blue!25}0.59{\textcolor{CtxTwoK}{(2k)}} / 0.48{\textcolor{CtxSixteenK}{(16k)}}
& \cellcolor{blue!25}0.67{\textcolor{CtxTwoK}{(2k)}} / 0.55{\textcolor{CtxEightK}{(8k)}} \\

& ClimateFever
& \cellcolor{blue!15}0.76{\textcolor{CtxSixteenK}{(16k)}} / 0.67{\textcolor{CtxFourK}{(4k)}}
& \cellcolor{blue!15}0.66{\textcolor{CtxTwoK}{(2k)}} / 0.57{\textcolor{CtxEightK}{(8k)}}
& \cellcolor{blue!15}0.70{\textcolor{CtxTwoK}{(2k)}} / 0.63{\textcolor{CtxFourK}{(4k)}}
& \cellcolor{blue!25}0.69{\textcolor{CtxTwoK}{(2k)}} / 0.53{\textcolor{CtxFourK}{(4k)}}
& \cellcolor{blue!15}0.65{\textcolor{CtxEightK}{(8k)}} / 0.60{\textcolor{CtxSixteenK}{(16k)}}
& \cellcolor{blue!8}0.62{\textcolor{CtxEightK}{(8k)}} / 0.58{\textcolor{CtxFourK}{(4k)}}
& \cellcolor{blue!25}0.65{\textcolor{CtxTwoK}{(2k)}} / 0.54{\textcolor{CtxFourK}{(4k)}}
& \cellcolor{blue!40}0.72{\textcolor{CtxTwoK}{(2k)}} / 0.47{\textcolor{CtxFourK}{(4k)}}
& \cellcolor{blue!40}0.67{\textcolor{CtxTwoK}{(2k)}} / 0.42{\textcolor{CtxFourK}{(4k)}}
& \cellcolor{blue!40}0.69{\textcolor{CtxTwoK}{(2k)}} / 0.46{\textcolor{CtxFourK}{(4k)}}
& \cellcolor{blue!25}0.72{\textcolor{CtxTwoK}{(2k)}} / 0.57{\textcolor{CtxSixteenK}{(16k)}} \\

\midrule

\multirow{3}{*}{Qwen2.5-7B} & Hover
& \cellcolor{blue!8}0.60{\textcolor{CtxFourK}{(4k)}} / 0.58{\textcolor{CtxSixteenK}{(16k)}}
& \cellcolor{blue!15}0.57{\textcolor{CtxTwoK}{(2k)}} / 0.52{\textcolor{CtxEightK}{(8k)}}
& \cellcolor{blue!15}0.56{\textcolor{CtxTwoK}{(2k)}} / 0.51{\textcolor{CtxSixteenK}{(16k)}}
& \cellcolor{blue!15}0.57{\textcolor{CtxTwoK}{(2k)}} / 0.51{\textcolor{CtxSixteenK}{(16k)}}
& \cellcolor{blue!15}0.56{\textcolor{CtxFourK}{(4k)}} / 0.51{\textcolor{CtxEightK}{(8k)}}
& \cellcolor{blue!15}0.56{\textcolor{CtxTwoK}{(2k)}} / 0.51{\textcolor{CtxSixteenK}{(16k)}}
& \cellcolor{blue!15}0.56{\textcolor{CtxTwoK}{(2k)}} / 0.51{\textcolor{CtxSixteenK}{(16k)}}
& \cellcolor{blue!15}0.57{\textcolor{CtxTwoK}{(2k)}} / 0.51{\textcolor{CtxSixteenK}{(16k)}}
& \cellcolor{blue!8}0.56{\textcolor{CtxTwoK}{(2k)}} / 0.52{\textcolor{CtxSixteenK}{(16k)}}
& \cellcolor{blue!8}0.56{\textcolor{CtxFourK}{(4k)}} / 0.53{\textcolor{CtxEightK}{(8k)}}
& \cellcolor{blue!8}0.57{\textcolor{CtxTwoK}{(2k)}} / 0.54{\textcolor{CtxSixteenK}{(16k)}} \\

& FEVEROUS
& \cellcolor{blue!15}0.75{\textcolor{CtxFourK}{(4k)}} / 0.69{\textcolor{CtxTwoK}{(2k)}}
& \cellcolor{blue!25}0.65{\textcolor{CtxTwoK}{(2k)}} / 0.52{\textcolor{CtxEightK}{(8k)}}
& \cellcolor{blue!15}0.61{\textcolor{CtxFourK}{(4k)}} / 0.52{\textcolor{CtxSixteenK}{(16k)}}
& \cellcolor{blue!25}0.66{\textcolor{CtxFourK}{(4k)}} / 0.52{\textcolor{CtxSixteenK}{(16k)}}
& \cellcolor{blue!25}0.65{\textcolor{CtxTwoK}{(2k)}} / 0.52{\textcolor{CtxSixteenK}{(16k)}}
& \cellcolor{blue!15}0.63{\textcolor{CtxTwoK}{(2k)}} / 0.52{\textcolor{CtxSixteenK}{(16k)}}
& \cellcolor{blue!25}0.64{\textcolor{CtxTwoK}{(2k)}} / 0.51{\textcolor{CtxSixteenK}{(16k)}}
& \cellcolor{blue!25}0.65{\textcolor{CtxTwoK}{(2k)}} / 0.51{\textcolor{CtxSixteenK}{(16k)}}
& \cellcolor{blue!15}0.63{\textcolor{CtxTwoK}{(2k)}} / 0.53{\textcolor{CtxSixteenK}{(16k)}}
& \cellcolor{blue!25}0.65{\textcolor{CtxFourK}{(4k)}} / 0.54{\textcolor{CtxEightK}{(8k)}}
& \cellcolor{blue!15}0.76{\textcolor{CtxFourK}{(4k)}} / 0.67{\textcolor{CtxEightK}{(8k)}} \\

& ClimateFever
& \cellcolor{blue!15}0.74{\textcolor{CtxFourK}{(4k)}} / 0.67{\textcolor{CtxTwoK}{(2k)}}
& \cellcolor{blue!40}0.55{\textcolor{CtxTwoK}{(2k)}} / 0.37{\textcolor{CtxEightK}{(8k)}}
& \cellcolor{blue!40}0.58{\textcolor{CtxTwoK}{(2k)}} / 0.33{\textcolor{CtxSixteenK}{(16k)}}
& \cellcolor{blue!40}0.55{\textcolor{CtxTwoK}{(2k)}} / 0.31{\textcolor{CtxSixteenK}{(16k)}}
& \cellcolor{blue!25}0.50{\textcolor{CtxTwoK}{(2k)}} / 0.33{\textcolor{CtxEightK}{(8k)}}
& \cellcolor{blue!25}0.49{\textcolor{CtxTwoK}{(2k)}} / 0.31{\textcolor{CtxSixteenK}{(16k)}}
& \cellcolor{blue!40}0.57{\textcolor{CtxTwoK}{(2k)}} / 0.31{\textcolor{CtxSixteenK}{(16k)}}
& \cellcolor{blue!40}0.54{\textcolor{CtxTwoK}{(2k)}} / 0.31{\textcolor{CtxSixteenK}{(16k)}}
& \cellcolor{blue!25}0.44{\textcolor{CtxTwoK}{(2k)}} / 0.33{\textcolor{CtxSixteenK}{(16k)}}
& \cellcolor{blue!25}0.52{\textcolor{CtxTwoK}{(2k)}} / 0.34{\textcolor{CtxSixteenK}{(16k)}}
& \cellcolor{blue!25}0.67{\textcolor{CtxFourK}{(4k)}} / 0.57{\textcolor{CtxSixteenK}{(16k)}} \\

\midrule

\multirow{3}{*}{Qwen3-8B} & Hover
& \cellcolor{blue!15}0.60{\textcolor{CtxFourK}{(4k)}} / 0.54{\textcolor{CtxEightK}{(8k)}}
& \cellcolor{blue!8}0.58{\textcolor{CtxTwoK}{(2k)}} / 0.55{\textcolor{CtxFourK}{(4k)}}
& \cellcolor{blue!8}0.58{\textcolor{CtxTwoK}{(2k)}} / 0.56{\textcolor{CtxFourK}{(4k)}}
& \cellcolor{blue!8}0.58{\textcolor{CtxTwoK}{(2k)}} / 0.55{\textcolor{CtxFourK}{(4k)}}
& \cellcolor{blue!15}0.58{\textcolor{CtxTwoK}{(2k)}} / 0.54{\textcolor{CtxFourK}{(4k)}}
& \cellcolor{blue!8}0.57{\textcolor{CtxTwoK}{(2k)}} / 0.54{\textcolor{CtxFourK}{(4k)}}
& \cellcolor{blue!8}0.57{\textcolor{CtxTwoK}{(2k)}} / 0.53{\textcolor{CtxFourK}{(4k)}}
& \cellcolor{blue!15}0.57{\textcolor{CtxTwoK}{(2k)}} / 0.51{\textcolor{CtxFourK}{(4k)}}
& \cellcolor{blue!15}0.58{\textcolor{CtxTwoK}{(2k)}} / 0.52{\textcolor{CtxFourK}{(4k)}}
& \cellcolor{blue!8}0.56{\textcolor{CtxTwoK}{(2k)}} / 0.52{\textcolor{CtxFourK}{(4k)}}
& \cellcolor{blue!8}0.57{\textcolor{CtxFourK}{(4k)}} / 0.55{\textcolor{CtxEightK}{(8k)}} \\

& FEVEROUS
& 0.80{\textcolor{CtxFourK}{(4k)}} / 0.79{\textcolor{CtxSixteenK}{(16k)}}
& \cellcolor{blue!25}0.72{\textcolor{CtxTwoK}{(2k)}} / 0.57{\textcolor{CtxFourK}{(4k)}}
& \cellcolor{blue!25}0.69{\textcolor{CtxEightK}{(8k)}} / 0.57{\textcolor{CtxFourK}{(4k)}}
& \cellcolor{blue!25}0.69{\textcolor{CtxTwoK}{(2k)}} / 0.56{\textcolor{CtxFourK}{(4k)}}
& \cellcolor{blue!25}0.70{\textcolor{CtxSixteenK}{(16k)}} / 0.55{\textcolor{CtxFourK}{(4k)}}
& \cellcolor{blue!25}0.69{\textcolor{CtxSixteenK}{(16k)}} / 0.55{\textcolor{CtxFourK}{(4k)}}
& \cellcolor{blue!25}0.66{\textcolor{CtxTwoK}{(2k)}} / 0.53{\textcolor{CtxFourK}{(4k)}}
& \cellcolor{blue!25}0.66{\textcolor{CtxTwoK}{(2k)}} / 0.53{\textcolor{CtxFourK}{(4k)}}
& \cellcolor{blue!25}0.66{\textcolor{CtxTwoK}{(2k)}} / 0.53{\textcolor{CtxFourK}{(4k)}}
& \cellcolor{blue!25}0.65{\textcolor{CtxTwoK}{(2k)}} / 0.53{\textcolor{CtxFourK}{(4k)}}
& \cellcolor{blue!15}0.66{\textcolor{CtxTwoK}{(2k)}} / 0.60{\textcolor{CtxFourK}{(4k)}} \\

& ClimateFever
& \cellcolor{blue!8}0.76{\textcolor{CtxEightK}{(8k)}} / 0.72{\textcolor{CtxTwoK}{(2k)}}
& \cellcolor{blue!40}0.62{\textcolor{CtxTwoK}{(2k)}} / 0.40{\textcolor{CtxFourK}{(4k)}}
& \cellcolor{blue!25}0.53{\textcolor{CtxTwoK}{(2k)}} / 0.42{\textcolor{CtxFourK}{(4k)}}
& \cellcolor{blue!40}0.56{\textcolor{CtxTwoK}{(2k)}} / 0.40{\textcolor{CtxFourK}{(4k)}}
& \cellcolor{blue!40}0.59{\textcolor{CtxSixteenK}{(16k)}} / 0.35{\textcolor{CtxFourK}{(4k)}}
& \cellcolor{blue!25}0.53{\textcolor{CtxSixteenK}{(16k)}} / 0.38{\textcolor{CtxFourK}{(4k)}}
& \cellcolor{blue!40}0.52{\textcolor{CtxTwoK}{(2k)}} / 0.33{\textcolor{CtxFourK}{(4k)}}
& \cellcolor{blue!25}0.49{\textcolor{CtxTwoK}{(2k)}} / 0.30{\textcolor{CtxFourK}{(4k)}}
& \cellcolor{blue!25}0.47{\textcolor{CtxTwoK}{(2k)}} / 0.33{\textcolor{CtxFourK}{(4k)}}
& \cellcolor{blue!25}0.48{\textcolor{CtxTwoK}{(2k)}} / 0.33{\textcolor{CtxFourK}{(4k)}}
& \cellcolor{blue!25}0.50{\textcolor{CtxEightK}{(8k)}} / 0.38{\textcolor{CtxFourK}{(4k)}} \\

\midrule
\midrule

\multirow{3}{*}{Qwen3-32B} & Hover
& \cellcolor{blue!15}0.59{\textcolor{CtxTwoK}{(2k)}} / 0.56{\textcolor{CtxEightK}{(8k)}}
& \cellcolor{blue!15}0.58{\textcolor{CtxTwoK}{(2k)}} / 0.54{\textcolor{CtxFourK}{(4k)}}
& \cellcolor{blue!8}0.58{\textcolor{CtxSixteenK}{(16k)}} / 0.55{\textcolor{CtxFourK}{(4k)}}
& \cellcolor{blue!15}0.58{\textcolor{CtxSixteenK}{(16k)}} / 0.54{\textcolor{CtxFourK}{(4k)}}
& \cellcolor{blue!8}0.57{\textcolor{CtxTwoK}{(2k)}} / 0.55{\textcolor{CtxFourK}{(4k)}}
& 0.57{\textcolor{CtxTwoK}{(2k)}} / 0.56{\textcolor{CtxFourK}{(4k)}}
& \cellcolor{blue!8}0.58{\textcolor{CtxSixteenK}{(16k)}} / 0.55{\textcolor{CtxFourK}{(4k)}}
& \cellcolor{blue!8}0.58{\textcolor{CtxSixteenK}{(16k)}} / 0.55{\textcolor{CtxEightK}{(8k)}}
& \cellcolor{blue!15}0.57{\textcolor{CtxSixteenK}{(16k)}} / 0.54{\textcolor{CtxFourK}{(4k)}}
& \cellcolor{blue!8}0.57{\textcolor{CtxTwoK}{(2k)}} / 0.55{\textcolor{CtxEightK}{(8k)}}
& \cellcolor{blue!15}0.57{\textcolor{CtxTwoK}{(2k)}} / 0.54{\textcolor{CtxEightK}{(8k)}} \\

& FEVEROUS
& \cellcolor{blue!8}0.84{\textcolor{CtxTwoK}{(2k)}} / 0.80{\textcolor{CtxFourK}{(4k)}}
& \cellcolor{blue!15}0.83{\textcolor{CtxSixteenK}{(16k)}} / 0.74{\textcolor{CtxFourK}{(4k)}}
& \cellcolor{blue!15}0.82{\textcolor{CtxSixteenK}{(16k)}} / 0.76{\textcolor{CtxFourK}{(4k)}}
& \cellcolor{blue!25}0.82{\textcolor{CtxSixteenK}{(16k)}} / 0.71{\textcolor{CtxFourK}{(4k)}}
& \cellcolor{blue!15}0.83{\textcolor{CtxSixteenK}{(16k)}} / 0.75{\textcolor{CtxFourK}{(4k)}}
& \cellcolor{blue!15}0.83{\textcolor{CtxSixteenK}{(16k)}} / 0.76{\textcolor{CtxFourK}{(4k)}}
& \cellcolor{blue!8}0.82{\textcolor{CtxEightK}{(8k)}} / 0.75{\textcolor{CtxFourK}{(4k)}}
& \cellcolor{blue!8}0.82{\textcolor{CtxSixteenK}{(16k)}} / 0.77{\textcolor{CtxFourK}{(4k)}}
& \cellcolor{blue!15}0.82{\textcolor{CtxSixteenK}{(16k)}} / 0.73{\textcolor{CtxFourK}{(4k)}}
& \cellcolor{blue!15}0.83{\textcolor{CtxSixteenK}{(16k)}} / 0.78{\textcolor{CtxFourK}{(4k)}}
& \cellcolor{blue!25}0.81{\textcolor{CtxSixteenK}{(16k)}} / 0.70{\textcolor{CtxFourK}{(4k)}} \\

& ClimateFever
& \cellcolor{blue!8}0.86{\textcolor{CtxFourK}{(4k)}} / 0.82{\textcolor{CtxSixteenK}{(16k)}}
& \cellcolor{blue!25}0.84{\textcolor{CtxFourK}{(4k)}} / 0.74{\textcolor{CtxSixteenK}{(16k)}}
& \cellcolor{blue!25}0.85{\textcolor{CtxTwoK}{(2k)}} / 0.72{\textcolor{CtxSixteenK}{(16k)}}
& \cellcolor{blue!25}0.82{\textcolor{CtxFourK}{(4k)}} / 0.68{\textcolor{CtxSixteenK}{(16k)}}
& \cellcolor{blue!15}0.79{\textcolor{CtxTwoK}{(2k)}} / 0.74{\textcolor{CtxSixteenK}{(16k)}}
& \cellcolor{blue!25}0.80{\textcolor{CtxTwoK}{(2k)}} / 0.67{\textcolor{CtxSixteenK}{(16k)}}
& \cellcolor{blue!40}0.82{\textcolor{CtxTwoK}{(2k)}} / 0.62{\textcolor{CtxSixteenK}{(16k)}}
& \cellcolor{blue!15}0.78{\textcolor{CtxEightK}{(8k)}} / 0.69{\textcolor{CtxSixteenK}{(16k)}}
& \cellcolor{blue!15}0.77{\textcolor{CtxEightK}{(8k)}} / 0.70{\textcolor{CtxSixteenK}{(16k)}}
& \cellcolor{blue!15}0.75{\textcolor{CtxEightK}{(8k)}} / 0.69{\textcolor{CtxSixteenK}{(16k)}}
& \cellcolor{blue!8}0.81{\textcolor{CtxEightK}{(8k)}} / 0.78{\textcolor{CtxTwoK}{(2k)}} \\

\midrule

\multirow{3}{*}{Llama-3.1-70B} & Hover
& 0.64{\textcolor{CtxSixteenK}{(16k)}} / 0.63{\textcolor{CtxTwoK}{(2k)}}
& \cellcolor{blue!8}0.63{\textcolor{CtxEightK}{(8k)}} / 0.61{\textcolor{CtxTwoK}{(2k)}}
& \cellcolor{blue!8}0.62{\textcolor{CtxFourK}{(4k)}} / 0.60{\textcolor{CtxTwoK}{(2k)}}
& 0.61{\textcolor{CtxEightK}{(8k)}} / 0.60{\textcolor{CtxSixteenK}{(16k)}}
& 0.61{\textcolor{CtxTwoK}{(2k)}} / 0.60{\textcolor{CtxFourK}{(4k)}}
& 0.61{\textcolor{CtxEightK}{(8k)}} / 0.60{\textcolor{CtxFourK}{(4k)}}
& 0.61{\textcolor{CtxEightK}{(8k)}} / 0.60{\textcolor{CtxFourK}{(4k)}}
& \cellcolor{blue!8}0.61{\textcolor{CtxEightK}{(8k)}} / 0.59{\textcolor{CtxFourK}{(4k)}}
& \cellcolor{blue!8}0.61{\textcolor{CtxEightK}{(8k)}} / 0.59{\textcolor{CtxFourK}{(4k)}}
& 0.60{\textcolor{CtxEightK}{(8k)}} / 0.59{\textcolor{CtxFourK}{(4k)}}
& \cellcolor{blue!8}0.60{\textcolor{CtxTwoK}{(2k)}} / 0.57{\textcolor{CtxSixteenK}{(16k)}} \\

& FEVEROUS
& 0.84{\textcolor{CtxTwoK}{(2k)}} / 0.83{\textcolor{CtxFourK}{(4k)}}
& 0.82{\textcolor{CtxEightK}{(8k)}} / 0.81{\textcolor{CtxSixteenK}{(16k)}}
& \cellcolor{blue!25}0.83{\textcolor{CtxFourK}{(4k)}} / 0.68{\textcolor{CtxSixteenK}{(16k)}}
& \cellcolor{blue!15}0.83{\textcolor{CtxTwoK}{(2k)}} / 0.75{\textcolor{CtxEightK}{(8k)}}
& \cellcolor{blue!15}0.83{\textcolor{CtxTwoK}{(2k)}} / 0.75{\textcolor{CtxEightK}{(8k)}}
& \cellcolor{blue!25}0.83{\textcolor{CtxTwoK}{(2k)}} / 0.71{\textcolor{CtxEightK}{(8k)}}
& \cellcolor{blue!25}0.82{\textcolor{CtxTwoK}{(2k)}} / 0.67{\textcolor{CtxEightK}{(8k)}}
& \cellcolor{blue!25}0.83{\textcolor{CtxTwoK}{(2k)}} / 0.64{\textcolor{CtxEightK}{(8k)}}
& \cellcolor{blue!25}0.83{\textcolor{CtxTwoK}{(2k)}} / 0.70{\textcolor{CtxEightK}{(8k)}}
& \cellcolor{blue!25}0.82{\textcolor{CtxTwoK}{(2k)}} / 0.71{\textcolor{CtxSixteenK}{(16k)}}
& \cellcolor{blue!8}0.82{\textcolor{CtxTwoK}{(2k)}} / 0.79{\textcolor{CtxFourK}{(4k)}} \\

& ClimateFever
& \cellcolor{blue!8}0.85{\textcolor{CtxTwoK}{(2k)}} / 0.83{\textcolor{CtxEightK}{(8k)}}
& \cellcolor{blue!25}0.83{\textcolor{CtxTwoK}{(2k)}} / 0.69{\textcolor{CtxSixteenK}{(16k)}}
& \cellcolor{blue!25}0.73{\textcolor{CtxTwoK}{(2k)}} / 0.54{\textcolor{CtxSixteenK}{(16k)}}
& \cellcolor{blue!40}0.79{\textcolor{CtxTwoK}{(2k)}} / 0.47{\textcolor{CtxSixteenK}{(16k)}}
& \cellcolor{blue!40}0.76{\textcolor{CtxTwoK}{(2k)}} / 0.50{\textcolor{CtxSixteenK}{(16k)}}
& \cellcolor{blue!40}0.73{\textcolor{CtxTwoK}{(2k)}} / 0.47{\textcolor{CtxSixteenK}{(16k)}}
& \cellcolor{blue!40}0.74{\textcolor{CtxTwoK}{(2k)}} / 0.41{\textcolor{CtxEightK}{(8k)}}
& \cellcolor{blue!40}0.72{\textcolor{CtxTwoK}{(2k)}} / 0.40{\textcolor{CtxEightK}{(8k)}}
& \cellcolor{blue!40}0.65{\textcolor{CtxTwoK}{(2k)}} / 0.44{\textcolor{CtxEightK}{(8k)}}
& \cellcolor{blue!40}0.74{\textcolor{CtxTwoK}{(2k)}} / 0.41{\textcolor{CtxFourK}{(4k)}}
& \cellcolor{blue!40}0.74{\textcolor{CtxTwoK}{(2k)}} / 0.54{\textcolor{CtxFourK}{(4k)}} \\

\bottomrule

\end{tabular}
\end{table*}
We finally analyze how the relative position of the evidence block affects verification accuracy (\ref{rq:position}). Figure~\ref{fig:combined_accuracy_5models} plots performance as a function of evidence depth (0--100\%) within the prompt.

Across models, accuracy is generally higher when evidence appears near the beginning (0\%) or end (100\%) of the prompt, and lower when placed in intermediate positions. This pattern is consistent with previous work findings \cite{liu-etal-2024-lost}, and confirms that evidence effectiveness depends on where it appears within the sequence.

Model sensitivity to positioning differs substantially. Llama-3.1-70B exhibits relatively stable performance across depths compared to its 8B counterpart, whose accuracy fluctuates more markedly when evidence is moved away from the edges. At the same time, both Qwen3 variants—particularly Qwen3-32B—display smoother accuracy curves across positions than Llama models, indicating lower positional sensitivity. Notably, Qwen3-32B remains consistently above the Claim Only baseline (light grey dotted line in Figure \ref{fig:combined_accuracy_5models}) across all evidence depths and context lengths, indicating that on average—as long as evidence is present—its verification performance improves regardless of where the evidence block is positioned within the prompt.

These differences point to potential architectural or training-related factors influencing how models distribute attention across long inputs. While scale improves robustness within a model family (e.g., 70B vs. 8B), cross-family comparisons suggest that design and training strategies may play an equally important role in mitigating positional bias.

Table~\ref{tab:position-breakdown} provides a complementary, position-wise quantification of these effects by reporting, for each evidence depth, the best and worst accuracy observed across context lengths for each model and dataset. The color shading encodes the magnitude of degradation ($\Delta=\text{Best}-\text{Worst}$), with darker cells indicating stronger sensitivity. This breakdown confirms that positional robustness varies not only across models but also across datasets. In particular, Qwen3-32B consistently exhibits smaller gaps between its best and worst context configurations across depths, reflected in lighter shading relative to other models, reinforcing the observation from Figure~\ref{fig:combined_accuracy_5models}.
At the same time, ClimateFEVER emerges as the most sensitive dataset to positional and contextual variation. Across multiple models and depths, the gap between best and worst configurations is substantially larger than for HOVER or FEVEROUS, as indicated by the darker cells in Table~\ref{tab:position-breakdown}. 

Taken together, these results show that where evidence is placed in the prompt directly impacts verification accuracy. Models differ in how effectively they use relevant information when it appears away from the beginning or end of the input. This highlights the practical importance of ordering evidence carefully in retrieval-augmented fact-checking systems.
\section{Conclusion and Future Work}
In this work, we investigated how LLMs perform in fact verification under varying contextual conditions. We showed that (i) models possess non-trivial parametric knowledge and can resolve a substantial portion of claims without external evidence (\ref{rq:pre}), (ii) when evidence is provided, increasing context length generally degrades verification accuracy on average (\ref{rq:len}), and (iii) evidence placed mid-context generally leads to lower accuracy (\ref{rq:position}). 

Across models, Qwen3-32B emerged as the most robust configuration, achieving the strongest overall performance and exhibiting comparatively stable behavior across context lengths and evidence depths.

These findings have practical implications for retrieval-augmented fact-checking systems. Simply increasing context length does not guarantee improved reasoning, and careful ordering of evidence within prompts can materially influence outcomes.

Future work should investigate the mechanisms underlying positional robustness, including architectural differences and training strategies for long-context utilization. Extending the analysis to retrieval pipelines with noisy or partially relevant evidence would further clarify how these effects manifest in realistic end-to-end fact-checking settings. Finally, controlled ablations on attention patterns and token-level attribution could provide deeper insight into how models allocate attention across long inputs.
%


\bibliographystyle{ACM-Reference-Format}
\bibliography{evidence_placement_factchecking}

\end{document}